\documentclass[11pt,a4paper]{article}
\usepackage[hyperref]{emnlp-ijcnlp-2019}
\usepackage{times}
\usepackage{latexsym}
\usepackage{url}
\usepackage[normalem]{ulem}
\usepackage{booktabs}
\usepackage{amsmath}
\usepackage{graphicx}
\usepackage{soul}
\usepackage{color}
\definecolor{lightgray}{rgb}{0.95,0.85,0.85}
\sethlcolor{lightgray}

\aclfinalcopy 

\title{GlossBERT: BERT for Word Sense Disambiguation \\
with Gloss Knowledge}

\author{Luyao Huang, Chi Sun, Xipeng Qiu\thanks{{ }{ }Corresponding author.}, Xuanjing Huang\\
    Shanghai Key Laboratory of Intelligent Information Processing, Fudan University\\
    School of Computer Science, Fudan University\\
    825 Zhangheng Road, Shanghai, China\\
    {\tt \{lyhuang18,sunc17,xpqiu,xjhuang\}@fudan.edu.cn} \\}

\date{}

\begin{document}
\maketitle

\begin{abstract}
Word Sense Disambiguation (WSD) aims to find the exact sense of an ambiguous word in a particular context. Traditional supervised methods rarely take into consideration the lexical resources like WordNet, which are widely utilized in knowledge-based methods. Recent studies have shown the effectiveness of incorporating gloss (sense definition) into neural networks for WSD. However, compared with traditional \textit{word expert} supervised methods, they have not achieved much improvement.
In this paper, we focus on how to better leverage gloss knowledge in a supervised neural WSD system. We construct \textit{context-gloss} pairs and propose three BERT-based models for WSD. We fine-tune the pre-trained BERT model on SemCor3.0 training corpus and the experimental results on several English all-words WSD benchmark datasets show that our approach outperforms the state-of-the-art systems \footnote{\url{https://github.com/HSLCY/GlossBERT}}.

\end{abstract}

\section{Introduction}
Word Sense Disambiguation (WSD) is a fundamental task and long-standing challenge in Natural Language Processing (NLP), which aims to find the exact sense of an ambiguous word in a particular context \citep{navigli2009word}. Previous WSD approaches can be grouped into two main categories: knowledge-based and supervised methods.

Knowledge-based WSD methods rely on lexical resources like WordNet \citep{miller1995wordnet} and usually exploit two kinds of lexical knowledge. The gloss, which defines a word sense meaning, is first utilized in Lesk algorithm \citep{lesk1986automatic} and then widely taken into account in many other approaches \citep{banerjee2002adapted, basile2014enhanced}. Besides, structural properties of semantic graphs are mainly used in graph-based algorithms \citep{agirre2014random, moro2014entity}.

Traditional supervised WSD methods \citep{zhong2010makes, shen2013coarse, iacobacci2016embeddings} focus on extracting manually designed features
and then train a dedicated classifier (\textit{word expert}) for every target lemma.

Although \textit{word expert} supervised WSD methods perform better, they are less flexible than knowledge-based methods in the all-words WSD task \citep{raganato2017neural}. Recent neural-based methods are devoted to dealing with this problem. \citet{kaageback2016word} present a supervised classifier based on Bi-LSTM, which shares parameters among all word types except the last layer. \citet{raganato2017neural} convert WSD task to a sequence labeling task, thus building a unified model for all polysemous words.
However, neither of them can totally beat the best \textit{word expert} supervised methods.

More recently, \citet{luo2018incorporating} propose to leverage the gloss information from WordNet and model the semantic relationship between the context and gloss in an improved memory network. Similarly, \citet{luo2018leveraging} introduce a (hierarchical) co-attention mechanism to generate co-dependent representations for the context and gloss. Their attempts prove that incorporating gloss knowledge into supervised WSD approach is helpful, but they still have not achieved much improvement, because they may not make full use of gloss knowledge.

\begin{table*}[t]
    \small
	\centering
	\begin{tabular}{l c c}
		\toprule
        \textbf{Sentence with four targets:}\\
        Your \uline{research} \uline{stopped} when a convenient \uline{assertion} could be \uline{made}. \\
        \\

		\textbf{Context-Gloss Pairs of the target word [research]} & Label & Sense Key \\
		\hline
		\texttt{[CLS]} Your research ... \texttt{[SEP]}	systematic investigation to ... \texttt{[SEP]} & Yes & research\%1:04:00:: \\
        \texttt{[CLS]} Your research ... \texttt{[SEP]} a search for knowledge \texttt{[SEP]} & No & research\%1:09:00::\\
		\texttt{[CLS]} Your research ... \texttt{[SEP]} inquire into \texttt{[SEP]} & No & research\%2:31:00::\\
		\texttt{[CLS]} Your research ... \texttt{[SEP]} attempt to find out in a ... \texttt{[SEP]} & No & research\%2:32:00::\\
		\\
		\textbf{Context-Gloss Pairs with weak supervision of the target word [research]} & Label & Sense Key \\
		\hline
		\texttt{[CLS]} Your \hl{``}research\hl{"} ... \texttt{[SEP]}	\hl{research:} systematic investigation to ... \texttt{[SEP]} & Yes & research\%1:04:00:: \\
        \texttt{[CLS]} Your \hl{``}research\hl{"} ... \texttt{[SEP]}	\hl{research:} a search for knowledge \texttt{[SEP]} & No & research\%1:09:00::\\
		\texttt{[CLS]} Your \hl{``}research\hl{"} ... \texttt{[SEP]}	\hl{research:} inquire into \texttt{[SEP]} & No & research\%2:31:00::\\
		\texttt{[CLS]} Your \hl{``}research\hl{"} ... \texttt{[SEP]}	\hl{research:} attempt to find out in a ... \texttt{[SEP]} & No & research\%2:32:00::\\
		\bottomrule
	\end{tabular}
\caption{\label{example} The construction methods. The sentence is taken from SemEval-2007 WSD dataset. The ellipsis ``..." indicates the remainder of the sentence or the gloss.}
\end{table*}

In this paper, we focus on how to better leverage gloss information in a supervised neural WSD system. Recently, the pre-trained language models, such as ELMo \citep{peters2018deep} and BERT \citep{devlin2018bert}, have shown their effectiveness to alleviate the effort of feature engineering. Especially, BERT has achieved excellent results in question answering (QA) and natural language inference (NLI). We construct \textit{context-gloss} pairs from glosses of all possible senses (in WordNet) of the target word, thus treating WSD task as a sentence-pair classification problem. We fine-tune the pre-trained BERT model on SemCor3.0 training corpus.

Recently\footnote{after we submitted the final version to the conference}, we are informed by \citet{vial2019sense}\footnote{their paper is available on arXiv after our first submission to the conference in May, 2019} that they also use BERT and incorporate WordNet as lexical knowledge in their supervised WSD system. But our work is much different from theirs. They exploit the semantic relationships between senses such as synonymy, hypernymy and hyponymy and rely on pre-trained BERT word vectors (feature-based approach); we leverage gloss knowledge (sense definition) and use BERT through fine-tuning procedures. Out of respect, we add their results in Table \ref{main results}. However, the results of their feature-based approach in the same experimental setup (single training set and single model) are not as good as our fine-tuning approach although their ensemble systems (with another training set WNGC) achieve better performance.

In particular, our contribution is two-fold:

1. We construct \textit{context-gloss} pairs and propose three BERT-based models for WSD.

2. We fine-tune the pre-trained BERT model on SemCor3.0 training corpus, and the experimental results on several English all-words WSD benchmark datasets show that our approach outperforms the state-of-the-art systems.

\section{Methodology}

In this section, we describe our method in detail.

\subsection{Task Definition}
In WSD, a sentence $s$ usually consists of a series of words: $\{w_1,\cdots,w_m\}$, and some of the words $\{w_{i_1},\cdots,w_{i_k}\}$ are targets $\{t_1,\cdots,t_k\}$ need to be disambiguated. For each target $t$, its candidate senses $\{c_1,\cdots,c_n\}$ come from entries of its lemma in a pre-defined sense inventory (usually WordNet). Therefore, WSD task aims to find the most suitable entry (symbolized as unique sense key) for each target in a sentence. See a sentence example in Table \ref{example}.

\subsection{BERT}
BERT \citep{devlin2018bert} is a new language representation model, and its architecture is a multi-layer bidirectional Transformer encoder. BERT model is pre-trained on a large corpus and two novel unsupervised prediction tasks, i.e., masked language model and next sentence prediction tasks are used in pre-training. When incorporating BERT into downstream tasks, the fine-tuning procedure is recommended. We fine-tune the pre-trained BERT model on WSD task.


\paragraph{BERT(Token-CLS)}
Since every target in a sentence needs to be disambiguated to find its exact sense, WSD task can be regarded as a token-level classification task. To incorporate BERT to WSD task, we take the final hidden state of the token corresponding to the target word (if more than one token, we average them) and add a classification layer for every target lemma, which is the same as the last layer of the Bi-LSTM model \citep{kaageback2016word}.

\subsection{GlossBERT}
BERT can explicitly model the relationship of a pair of texts, which
has shown to be beneficial to many pair-wise natural language understanding tasks. In order to fully leverage gloss information, we propose GlossBERT to construct \textit{context-gloss} pairs from all possible senses of the target word in WordNet, thus treating WSD task as a sentence-pair classification problem.

We describe our construction method with an example (See Table \ref{example}). There are four targets in this sentence, and here we take target word \textit{research} as an example:

\paragraph{Context-Gloss Pairs}
The sentence containing target words is denoted as \textit{context} sentence. For each target word, we extract glosses of all $N$ possible senses (here $N=4$) of the target word (research) in WordNet to obtain the \textit{gloss} sentence. \texttt{[CLS]} and \texttt{[SEP]} marks are added to the \textit{context-gloss} pairs to make it suitable for the input of BERT model. A similar idea is also used in aspect-based sentiment analysis \cite{sun2019utilizing}.

\paragraph{Context-Gloss Pairs with Weak Supervision}
Based on the previous construction method, we add weak supervised signals to the \textit{context-gloss} pairs (see the highlighted part in Table \ref{example}). The signal in the \textit{gloss} sentence aims to point out the target word, and the signal in the \textit{context} sentence aims to emphasize the target word considering the situation that a target word may occur more than one time in the same sentence.

Therefore, each target word has $N$ \textit{context-gloss} pair training instances ($label\in\{yes, no\} $). When testing, we output the probability of $label=yes$ of each  \textit{context-gloss} pair and choose the sense corresponding to the highest probability as the prediction label of the target word. We experiment with three GlossBERT models:

\paragraph{GlossBERT(Token-CLS)}
We use context-gloss pairs as input. We highlight the target word by taking the final hidden state of the token corresponding to the target word (if more than one token, we average them) and add a classification layer ($label\in\{yes, no\} $).

\paragraph{GlossBERT(Sent-CLS)}
We use context-gloss pairs as input. We take the final hidden state of the first token \texttt{[CLS]} as the representation of the whole sequence and add a classification layer ($label\in\{yes, no\} $), which does not highlight the target word.

\paragraph{GlossBERT(Sent-CLS-WS)}
We use context-gloss pairs with weak supervision as input. We take the final hidden state of the first token \texttt{[CLS]} and add a classification layer ($label\in\{yes, no\} $), which weekly highlight the target word by the weak supervision.

\section{Experiments}

\subsection{Datasets}

\begin{table}[t]
    \footnotesize
    \centering
    \begin{tabular}{|c|c|c|c|c|c|c|}
         \hline
         Dataset & Total & Noun & Verb & Adj & Adv \\
         \hline
         SemCor & 226036 & 87002 & 88334 & 31753 & 18947 \\
         \hline
         SE2 & 2282 & 1066 & 517 & 445 & 254 \\
         \hline
         SE3 & 1850 & 900 & 588 & 350 & 12  \\
         \hline
         SE07 & 455 & 159 & 296 & 0 & 0 \\
         \hline
         SE13 & 1644 & 1644 & 0 & 0 & 0  \\
         \hline
         SE15 & 1022 & 531 & 251 & 160 & 80 \\
         \hline
    \end{tabular}
    \caption{Statistics of the different parts of speech annotations in English all-words WSD datasets. }
    \label{statistics}
\end{table}

The statistics of the WSD datasets are shown in Table \ref{statistics}.

\begin{table*} [t]
\small
\centering

\resizebox{0.99\linewidth}{!}
{
\begin{tabular}{|l|c|c c c c|c c c c|c|}
    \hline
    & \multicolumn{1}{c|}{Dev} &
    \multicolumn{4}{c|}{Test Datasets} &
    \multicolumn{5}{c|}{Concatenation of Test Datasets} \\
    \hline
    System & SE07 & SE2 & SE3 & SE13 & SE15 & Noun & Verb & Adj & Adv & \bf All \\
    \hline
    MFS baseline & 54.5 & 65.6 & 66.0 & 63.8 & 67.1 & 67.7 & 49.8 & 73.1 & 80.5 & 65.5 \\
    \hline
    Lesk$_{ext+emb}$  & 56.7 & 63.0 & 63.7 & 66.2 & 64.6 & 70.0 & 51.1 & 51.7 & 80.6 & 64.2 \\
    Babelfy & 51.6 & 67.0 & 63.5 & 66.4 & 70.3 & 68.9 & 50.7 & 73.2 & 79.8 & 66.4 \\
    \hline
    IMS & 61.3 & 70.9 & 69.3 & 65.3 & 69.5 & 70.5 & 55.8 & 75.6 & 82.9 & 68.9 \\
    IMS$_{+emb}$ & 62.6 & 72.2 & 70.4 & 65.9 & 71.5 & 71.9 & 56.6 & 75.9 & 84.7 & 70.1 \\
    \hline
    Bi-LSTM & - & 71.1 & 68.4 & 64.8 & 68.3 & 69.5 & 55.9 & 76.2 & 82.4 & 68.4 \\
    Bi-LSTM$_{+ att. + LEX + POS}$  & 64.8 & 72.0 & 69.1 & 66.9 & 71.5 &71.5 & 57.5 & 75.0 & 83.8 & 69.9 \\
    GAS$_{ext}$ (Linear) & - &  72.4 & 70.1 & 67.1 & 72.1 & 71.9 &  58.1 &  76.4 &  84.7 & 70.4\\
    GAS$_{ext}$ (Concatenation)& - & 72.2 & 70.5 & 67.2 & 72.6 & 72.2 & 57.7 &   76.6 &   85.0 & 70.6\\
    CAN$^s$  & - & 72.2 & 70.2 & 69.1 & 72.2 & 73.5 & 56.5 & 76.6 & 80.3 & 70.9 \\
    HCAN & - & 72.8 & 70.3 & 68.5 & 72.8 & 72.7 & 58.2 & 77.4 & 84.1 & 71.1 \\
    \hline
    SemCor, hypernyms (single) & - & - & - & - & - & - & - & - & - & 75.6 \\
    SemCor, hypernyms (ensemble)$\dag$ & 69.5 & 77.5 & 77.4 & 76.0 & 78.3 & 79.6 & 65.9 & 79.5 & 85.5 & 76.7 \\ 
    SemCor+WNGC, hypernyms (single)$\ddag$ & - & - & - & - & - & - & - & - & - & 77.1 \\
    SemCor+WNGC, hypernyms (ensemble)$\dag$ $\ddag$ & 73.4 & 79.7 & 77.8 & 78.7 & 82.6 & 81.4 & 68.7 & 83.7 & 85.5 & 79.0 \\
    \hline
    BERT(Token-CLS) & 61.1 & 69.7 & 69.4 & 65.8 & 69.5 & 70.5 & 57.1 & 71.6 & 83.5 & 68.6 \\
    GlossBERT(Sent-CLS) & 69.2 & 76.5 & 73.4 & 75.1 & 79.5 & 78.3 & 64.8 & 77.6 & 83.8 & 75.8 \\
    GlossBERT(Token-CLS) & 71.9 & 77.0 & \bf 75.4 & 74.6 & 79.3 & 78.3 & 66.5 & \bf 78.6 & 84.4 & 76.3 \\
    GlossBERT(Sent-CLS-WS) & \bf 72.5 & \bf 77.7 & 75.2 & \bf 76.1 & \bf 80.4 & \bf 79.3 & \bf 66.9 & 78.2 & \bf 86.4 & \bf 77.0 \\
    \hline
    \end{tabular}
    }
    \caption{F1-score (\%) for fine-grained English all-words WSD on the test sets in the framework of \citet{raganato2017word} (including the development set \textbf{SE07}). The six blocks list the MFS baseline, two knowledge-based systems, two traditional \textit{word expert} supervised systems, six recent neural-based systems, one BERT feature-based system and our systems, respectively. Results in first three blocks come from \citet{raganato2017word}, and others from the corresponding papers. $\dag$ values are ensemble systems and $\ddag$ values are models trained on both SemCor and WNGC. \textbf{Bold} font indicates best single model system trained on SemCor, i.e. excludes $\dag$ $\ddag$ values since it is meaningless to compare ensemble systems and models trained on two training sets with our single model trained on SemCor training set only.
    } \label{main results}.
    \vspace{-0.2in}
\end{table*}

\paragraph{Training Dataset}
Following previous work \citep{luo2018leveraging,luo2018incorporating,raganato2017neural,raganato2017word,iacobacci2016embeddings,zhong2010makes}, we choose SemCor3.0 as training corpus, which is the largest corpus manually annotated with WordNet sense for WSD.

\paragraph{Evaluation Datasets}
We evaluate our method on several English all-words WSD datasets. For a fair comparison, we use the benchmark datasets proposed by \citet{raganato2017word} which include five standard all-words fine-grained WSD datasets from the Senseval and SemEval competitions: Senseval-2 (\textbf{SE2}), Senseval-3 (\textbf{SE3}), SemEval-2007 (\textbf{SE07}), SemEval-2013 (\textbf{SE13}) and SemEval-2015 (\textbf{SE15}). Following \citet{luo2018leveraging}, \citet{luo2018incorporating} and \citet{raganato2017neural}, we choose \textbf{SE07}, the smallest among these test sets, as the development set.

\paragraph{WordNet}
Since \citet{raganato2017word} map all the sense annotations in these datasets from their original versions to WordNet 3.0, we extract word sense glosses from WordNet 3.0.

\subsection{Settings}
We use the pre-trained uncased BERT$_\mathrm{BASE}$ model\footnote{https://storage.googleapis.com/bert\_models/2018\_10\_18/\\uncased\_L-12\_H-768\_A-12.zip} for fine-tuning, because we find that BERT$_\mathrm{LARGE}$ model performs slightly worse than BERT$_\mathrm{BASE}$ in this task. The number of Transformer blocks is 12, the number of the hidden layer is 768, the number of self-attention heads is 12, and the total number of parameters of the pre-trained model is 110M. When fine-tuning, we use the development set (\textbf{SE07}) to find the optimal settings for our experiments. We keep the dropout probability at 0.1, set the number of epochs to 4. The initial learning rate is 2e-5, and the batch size is 64.

\subsection{Results}

Table \ref{main results} shows the performance of our method on the English all-words WSD benchmark datasets. We compare our approach with previous methods.

The first block shows the MFS baseline, which selects the most frequent sense in the training corpus for each target word.

The second block shows two knowledge-based systems. Lesk$_{ext+emb}$ \citep{basile2014enhanced} is a variant of Lesk algorithm \citep{lesk1986automatic} by calculating the gloss-context overlap of the target word. Babelfy \citep{moro2014entity} is a unified graph-based approach which exploits the semantic network structure from BabelNet.

The third block shows two \textit{word expert} traditional supervised systems. IMS \citep{zhong2010makes} is a flexible framework which trains SVM classifiers and uses local features. And IMS$_{+emb}$ \citep{iacobacci2016embeddings} is the best configuration of the IMS framework, which also integrates word embeddings as features.


The fourth block shows several recent neural-based methods. Bi-LSTM \citep{kaageback2016word} is a baseline for neural models. Bi-LSTM$_{+ att. + LEX + POS}$ \citep{raganato2017neural} is a multi-task learning framework for WSD, POS tagging, and LEX with self-attention mechanism, which converts WSD to a sequence learning task. GAS$_{ext}$ \citep{luo2018incorporating} is a variant of GAS which is a gloss-augmented variant of the memory network by extending gloss knowledge. CAN$^s$ and HCAN \citep{luo2018leveraging} are sentence-level and hierarchical co-attention neural network models which leverage gloss knowledge.

The fifth block are feature-based BERT models \cite{vial2019sense} which exploit the semantic relationships between senses such as synonymy, hypernymy and hyponymy, and use pre-trained BERT embeddings and transformer encoder layers. It is worth noting that our fine-tuning method is superior to their feature-based method under the same experimental setup (single model + SemCor training set). 

In the last block, we report the performance of our method. BERT(Token-CLS) is our baseline, which does not incorporate gloss information, and it performs slightly worse than previous traditional supervised methods and recent neural-based methods. It proves that directly using BERT cannot obtain performance growth. The other three methods outperform other models by a substantial margin, which proves that the improvements come from leveraging BERT to better exploit gloss information. It is worth noting that our method achieves significant improvements in \textbf{SE07} and \textbf{Verb} over previous methods, which have the highest ambiguity level among all datasets and all POS tags respectively according to \citet{raganato2017word}.

Moreover, GlossBERT(Token-CLS) performs better than GlossBERT(Sent-CLS), which proves that highlighting the target word in the sentence is important. However, the weakly highlighting method GlossBERT(Sent-CLS-WS) performs best in most circumstances, which may result from its combination of the advantages of the other two methods.

\subsection{Discussion}
There are two main reasons for the great improvements of our experimental results. First, we construct \textit{context-gloss} pairs and convert WSD problem to a sentence-pair classification task which is similar to NLI tasks and train only one classifier, which is equivalent to expanding the corpus. Second, we leverage BERT \citep{devlin2018bert} to better exploit the gloss information. BERT model shows its advantage in dealing with sentence-pair classification tasks by its amazing improvement on QA and NLI tasks. This advantage comes from both of its two novel unsupervised prediction tasks.

Compared with traditional \textit{word expert} supervised methods, our GlossBERT shows its effectiveness to alleviate the effort of feature engineering and does not require training a dedicated classifier for every target lemma. 
Compared with recent neural-based methods, our solution is more intuitive and can make better use of gloss knowledge. Besides, our approach demonstrates that when we fine-tune BERT on a downstream task, converting it into a sentence-pair classification task may be a good choice.

\section{Conclusion}
In this paper, we seek to better leverage gloss knowledge in a supervised neural WSD system. We propose a new solution to WSD by constructing \textit{context-gloss} pairs and then converting WSD to a sentence-pair classification task. We fine-tune the pre-trained BERT model on SemCor3.0 training corpus, and the experimental results on several English all-words WSD benchmark datasets show that our approach outperforms the state-of-the-art systems.



\section*{Acknowledgments}
We would like to thank the anonymous reviewers for their valuable comments. The research work is supported by National Natural Science Foundation of China (No. 61751201 and 61672162),
Shanghai Municipal Science and Technology Commission (16JC1420401 and 17JC1404100), Shanghai Municipal Science and Technology Major Project (No.2018SHZDZX01) and ZJLab.

\bibliography{WSD}
\bibliographystyle{acl_natbib}

\end{document}